# Enhancing Natural Language Inference Performance with Knowledge Graph for COVID-19 Automated Fact-Checking in Indonesian Language[#]

**Arief Purnama Muharram[1,*] & Ayu Purwarianti[1,2]**

[1]School of Electrical Engineering and Informatics, Institut Teknologi Bandung, Jalan Ganesa No. 10, Bandung 40132, Indonesia
[2]Center for Artificial Intelligence (U-COE AI-VLB), Institut Teknologi Bandung, Jalan Ganesa No. 10 Bandung 40132, Indonesia
[*]E-mail: ariefpurnamamuharram@gmail.com

**Abstract.** Automated fact-checking is a key strategy to overcome the spread of COVID-19 misinformation on the internet. These systems typically leverage deep learning approaches through natural language inference (NLI) to verify the truthfulness of information based on supporting evidence. However, one challenge that arises in deep learning is performance stagnation due to a lack of knowledge during training. This study proposes using a knowledge graph (KG) as external knowledge to enhance NLI performance for automated COVID-19 fact-checking in the Indonesian language. The proposed model architecture comprises three modules: a fact module, an NLI module, and a classifier module. The fact module processes information from the KG, while the NLI module handles semantic relationships between the given premise and hypothesis. The representation vectors from both modules are concatenated and fed into the classifier module to produce the final result. The model was trained using the generated Indonesian COVID-19 fact-checking dataset and the COVID-19 KG Bahasa Indonesia. Our study demonstrates that incorporating KGs can significantly improve NLI performance in fact-checking, achieving a maximum accuracy of 0.8616. This suggests that KGs are a valuable component for enhancing NLI performance in automated fact-checking.

## 1 Introduction

COVID-19, also known as Coronavirus Disease 2019, is an acute inflammatory disease caused by SARS-CoV-2 that affects the human respiratory system. The signs and symptoms of COVID-19 include cough, fever, and shortness of breath. COVID-19 was first announced in late 2019 and has since become a worldwide pandemic. At that time, COVID-19 became the main global health concern due to its high contagiousness and the mortality rate it caused, with efforts to find a treatment still in progress. Therefore, every country was forced to formulate an effective strategy to overcome the pandemic [1]. One of the strategies at the public health level was to ensure that people received accurate information. In such conditions, accurate information can help people understand the current situation, and therefore, proper action can be taken [2].

---

[#]Accepted for publication in the Journal of ICT Research and Applications (JICTRA)

With the advancement of the internet, people now tend to seek information online, including health-related information [3]. Online news portals and social media have become popular places for seeking such information [4]. This trend has brought advantages for people in finding reliable information faster. Furthermore, a study by Manika et al. [5] revealed that exposure to reliable online health information has had a positive impact on health-related behavior changes. This confirms the advantage of seeking health information online. However, despite this, this information-seeking behavior trend has also made people vulnerable to receiving misinformation [6].

Misinformation is simply defined as information that contradicts the facts [7]. Another definition of misinformation refers to information that is 'explicitly false' compared to what has been determined or believed by expert consensus [8]. Misinformation cannot be neglected, as it can have serious consequences, especially in the context of public health [2,5,8]. Misinformation can create distrust among people towards public health efforts, leading to failures in combating certain public health-related problems [2]. For example, misinformation about the COVID-19 vaccine has built negative sentiments in the public towards the vaccine [9], leading to lower adoption among the population. The widespread dissemination of misinformation through the internet can be explained by the abundance of unvalidated information spread through online channels, such as social media and news portals [6]. Therefore, attention must be given to overcoming this issue. One of the solutions is verifying the truthfulness of information through a process known as fact-checking [10,11].

Fact-checking is a journalistic process to verify the truthfulness of information [12]. At the beginning, fact-checking is a human labor- and time-intensive process [12] involving collecting supporting evidence and verifying the truthfulness of information according to the collected and supported evidence [11]. However, with the abundance of user-generated content on the internet, it is almost impossible to do it manually [11,13]. Thanks to the advancement of artificial intelligence and natural language processing, the fact-checking process paradigm has shifted towards automated fact-checking systems [11].

An automated fact-checking system leverages the power of deep learning [11], usually involving natural language inference (NLI) [11,14,15] using existing pre-trained language models (PLMs), to verify the truthfulness of information based on collected supporting evidence. NLI can be simply defined as a task of determining the relationship between a premise sentence and a hypothesis sentence [16,17,18], where, in the context of fact-checking, the hypothesis is the information being verified (claim) and the premise is the supporting evidence. The resulting relationships can be entailment (fact), contradiction (misinformation), or neutral (cannot be determined) (Table 1).

**Table 1** Examples of NLI.

| Premise | Hypothesis | Label |
|---|---|---|
| Countries are advised to administer a third shot of the Sinopharm COVID-19 vaccine to protect seniors. | In an effort to protect people aged 60 years and over, a third dose of the Sinopharm vaccine is recommended. | Entailment |
| The PCR test process for detecting the virus involves duplicating the genetic RNA in the body. | The PCR test step to detect the virus does not involve the amplification of RNA genetic material. | Contradiction |
| Pregnant women who contract COVID-19 are at high risk of giving birth to stillborn or premature babies. | The risk of complications in babies increases if pregnant women are infected with COVID-19. | Neutral |

The use of NLI for fact-checking has the advantage of better results compared to the traditional approach due to its ability to perform complex computations without relying on hand-crafted features [11]. Meanwhile, the use of existing PLMs through the fine-tuning process offers the advantage of using pre-trained representations, thereby eliminating the need to train from scratch [19,20]. However, despite its superiority, one challenge that arises with the use of such deep learning models for fact-checking is performance stagnancy. This stagnation can possibly be explained by a lack of certain knowledge during the training phase [21]. This knowledge is important in terms of fact-checking, as the truthfulness of information often relies on the current knowledge, which is anchored to the time when the knowledge was created [22]. To overcome this issue, there is research interest in injecting external knowledge into the model to enhance its performance using a knowledge graph (KG) [23,24].

A knowledge graph is a directed graph that represents real-world knowledge [25]. The structure of a KG consists of nodes and edges, where nodes represent real-world objects and edges represent the relationships between them (Figure 1). Information in a KG is therefore often represented as triplets (node–edge–node) [26]. A KG can be an alternative for storing real-world knowledge or information. Among types of KG, domain-specific KGs are smaller in size but more reliable for domain-specific purposes (such as fact-checking) [25]. The COVID-19 KG

Bahasa Indonesia is an example of a domain-specific KG that contains information about COVID-19 represented using semantics in the Indonesian language [27].

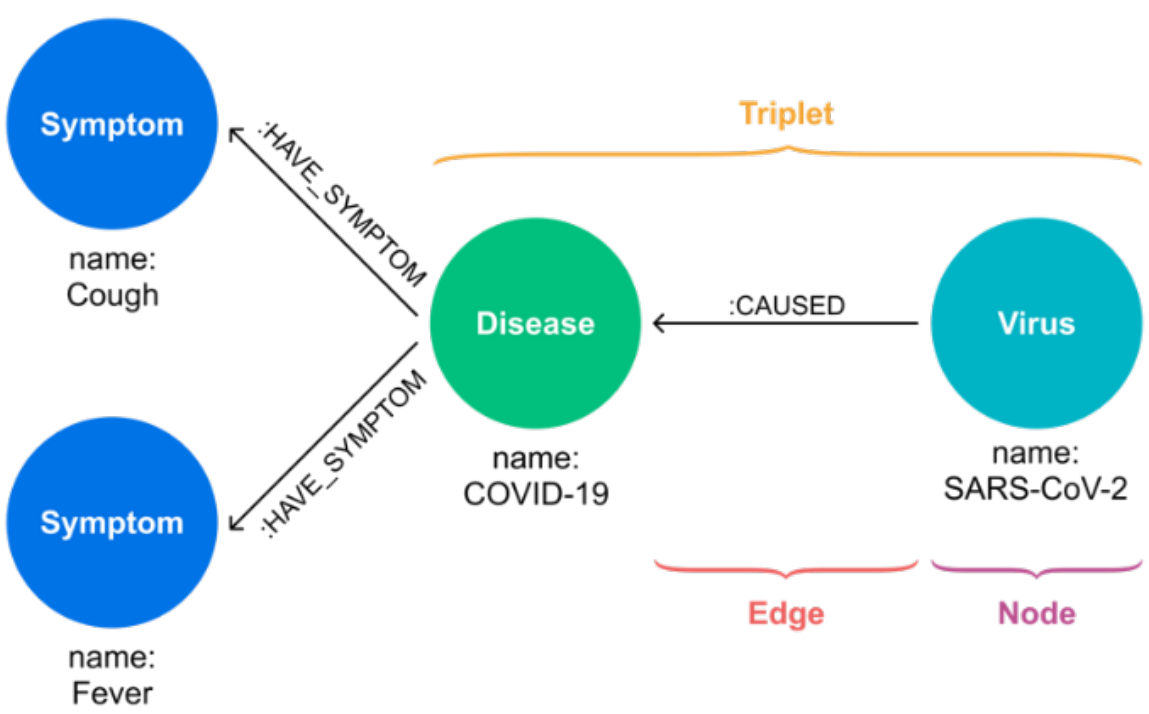

**Figure 1** Structure of a KG.

Given the potential of KGs to store real-world knowledge, this study proposes using a KG to enhance NLI performance for automated COVID-19 fact-checking in the Indonesian language. The role of the KG was to serve as external knowledge during the training and inference phases of the model. We selected the Indonesian language for our case study because it is considered a low-resource language [28] and it is used by around 270 million people. The key contributions of this study are summarized as follows:

1. We created an Indonesian Language COVID-19 fact-checking dataset comprised of 18,750 paired premise-hypothesis sentences divided into 3 labels (entailment, contradiction, neutral).
2. We propose a model architecture that can employ NLI and KG for fact-checking.
3. We conducted experiments with monolingual and multilingual pre-trained language models to evaluate our proposed deep learning architecture across various language models.

The rest of this paper is structured as follows: 1) Relevant Works: we describe works that are relevant to our study; 2) Methodology: we outline our proposed model architecture, dataset generation, and experimental procedures; 3) Results and Discussion: we present and discuss the experimental results; and 4) Conclusion: we summarize the findings of our study.

## 2 Relevant Works

Injecting external knowledge into a model through KGs is still a fascinating open research question. Many researchers are conducting studies to find the optimal method (both in terms of performance and the resulting complexity) to inject external knowledge into a model. To simplify, Yang et al. have further categorized these methods into six categories: feature-fused, embedding-combined, knowledge-supervised, data-structure unified, retrieval-based, and rule-guided [24]. Among these, data-structure unified, embedding-combined, and retrieval-based methods have our specific interest.

One challenge in injecting knowledge from KGs arises from the nature of KGs, which are represented as graphs. Therefore, the main idea behind a data-structure unified method is to transform and unify the input format into a defined, standardized structure. This unified data structure can then be used for downstream tasks [24]. K-BERT [29] is a well-known architecture that employs this method. The advantage of this approach is that it standardizes the input format. However, the main drawback is the increased complexity of input processing, which can lead to reduced performance if not properly handled.

In contrast, embedding-combined methods take advantage of embedding representations. The idea behind this approach is to encode the input from the KG through a representation learning module and then fuse the resulting representations with the token representations from the main input. This fused representation can then be used for downstream tasks [24]. KnowBERT [30] is known to use this method, which allows models to gain knowledge through the provided representation embeddings.

Another method of injecting knowledge is the retrieval-based method. This approach involves retrieving, selecting, and encoding the most relevant knowledge from extensive KG sources. Advantages of this method lie in its interpretability and practical application of knowledge [24]. KT-NET is one example of this method in use [31].

# 3 Methodology

## 3.1 Model Architecture

We approached the integration of knowledge from KG into models from a different perspective. In this study, we propose a model architecture that leverages the strengths of both embedding-based and retrieval-based methods. From the embedding-based method, we adopted the key concept of using fused embedding representations as input for downstream tasks. Meanwhile, from the retrieval-based method, we incorporated the concept of retrieving and selecting as much relevant information from the KG as possible, enabling the model to access extensive knowledge. Our approach allowed for straightforward knowledge integration while maintaining the simplicity of the model architecture. Figure 2 illustrates our proposed model architecture.

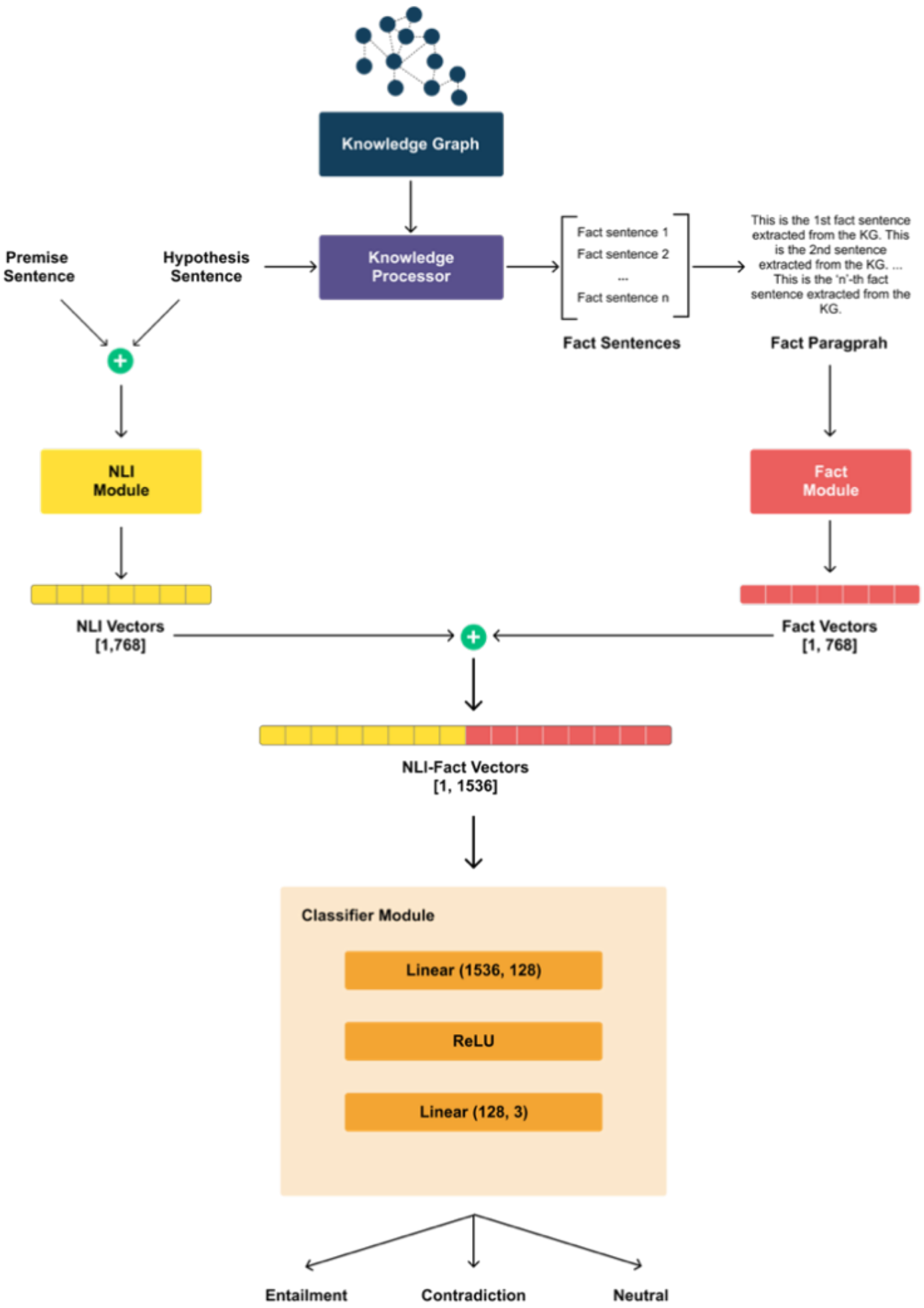

**Figure 2** Our proposed model architecture.

Our proposed model architecture consists of three modules: the NLI module, the fact module, and the classifier module. The NLI module is responsible for processing the semantic relationship between the given premise and hypothesis sentence, while the fact module handles the information from the fact paragraph. The resulting representation vectors from both modules are then fused (concatenated) into a single vector, which serves as the input for the classifier module. The classifier module then produces the final output (entailment, contradiction, or neutral). Both the NLI and fact modules are essentially PLMs, while the classifier module is a multi-layer perceptron network.

We define a 'fact paragraph' as a collection of 'fact sentences' combined to form a single paragraph. Each fact sentence is derived from a triplet retrieved from a KG, represented as $\{e^s, r, e^t\}$, where $e^s$ and $e^t$ represent the source and target entities (nodes), respectively, and $r$ represents the relationship between them. These elements are combined to form a single sentence. For example, given the triplet {'COVID-19', 'HAVE_SYMPTOM', 'cough'}, the fact sentence would be 'COVID-19 have symptom cough.' Figure 3 illustrates this straightforward process.

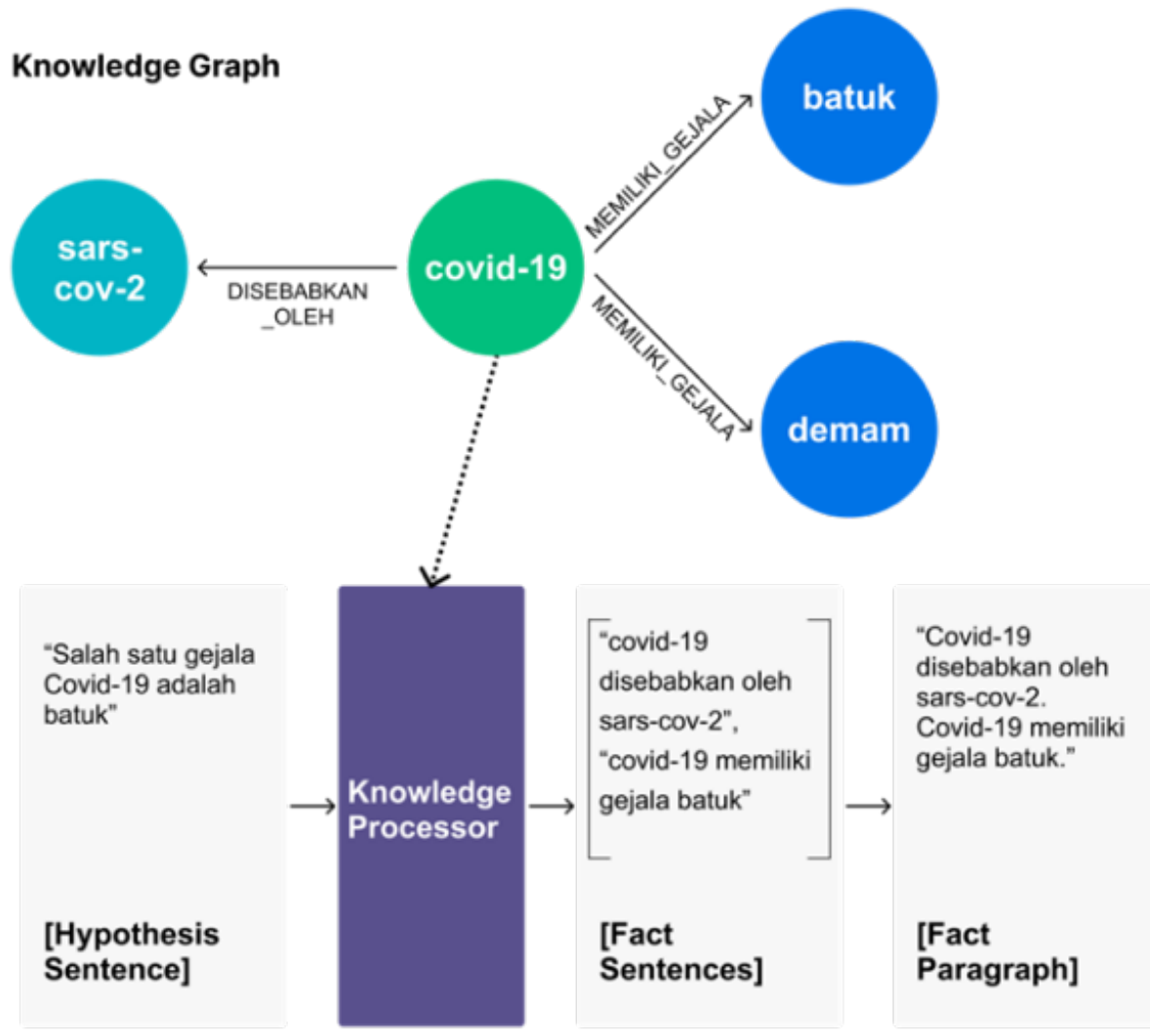

**Figure 3** Fact sentences and fact paragraph processing workflow.

To generate a fact sentence from the retrieved triplet, we used a word-matching retrieval mechanism approach. This mechanism is implemented in the knowledge processor part of the model. Given a knowledge graph (KG) as the source of external knowledge and a hypothesis sentence as the input query to retrieve the relevant triplet, the mechanism steps are as follows (Figure 3, Table 2):

1. The input sentence is split into words using a certain delimiter (in this case, white space). Words considered as stop words are removed. The stop words list used in this study was for the Indonesian language [32].
2. Each resulting word is then used as a query to find matched entities $e^s$ in the KG.

3. Each matched entity $e^s$ is then used to find the corresponding entity $e^t$ and its relationship $r$, forming a triplet $\{e^s, r, e^t\}$.
4. Each retrieved triplet is then joined together to form a fact sentence.
5. Lastly, each formed fact sentence is joined together to form a fact paragraph.

**Table 2** Data at each step of fact sentence and fact paragraph generation.

| Step | Data |
|---|---|
| Input | *'Salah satu gejala Covid-19 adalah batuk'* |
| 1-2 | (*'salah'*, *'satu'*, *'gejala'*, *'covid-19'*, *'batuk'*) |
| 3 | [(*'covid-19'*, *'DISEBABKAN_OLEH'*, *'sars-cov-2'*), (*'covid-19'*, *'MEMILIKI_GEJALA'*, *'batuk'*)] |
| 4 | [*'covid-19 disebabkan oleh sars-cov-2'*, *'covid-19 memiliki gejala batuk'*] |
| 5 | *'covid-19 disebabkan oleh sars-cov-2. Covid-19 memiliki gejala batuk.'* |

## 3.2 Dataset Generation

A dataset is needed to train and evaluate the model. In this case, we require a COVID-19 fact-checking dataset in the Indonesian language. To the best of our knowledge, there are currently no COVID-19 fact-checking or general fact-checking datasets available in Indonesian. Therefore, in this study, we generated our own fact-checking dataset with the help of ChatGPT. Specifically, we used ChatGPT 3.5 Turbo to create our synthetic dataset. ChatGPT has been proven in many studies to be capable of generating high-quality synthetic datasets for various downstream tasks at a lower cost [33,34,35]. Moreover, using generative large language models (LLMs) such as ChatGPT to generate synthetic datasets offers several advantages. It results in diverse and rich contextual datasets, which can lead to improved model performance [36]. However, these advantages also come with limitations that need to be considered when using LLMs to generate synthetic datasets. These limitations include the quality of the generated data, which depends on the training dataset and the model used; difficulties in handling niche domains such as the medical field due to limited exposure during training; and challenges in ensuring the semantic consistency, uniqueness, and correctness of the generated data [37]. Therefore, quality evaluation of the generated dataset is necessary. Figure 4 illustrates our dataset generation workflow in detail.

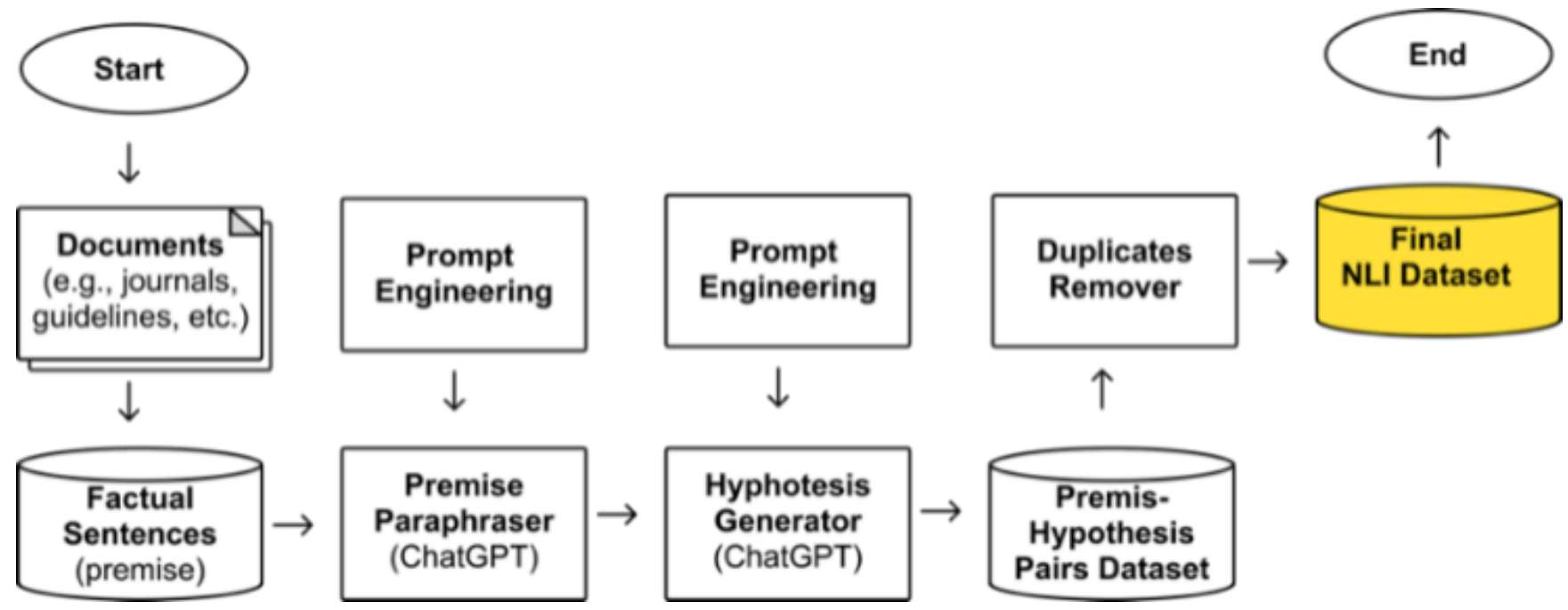

**Figure 4** Dataset generation workflow.

The process began by collecting factual sentences related to COVID-19 in the Indonesian language. These sentences were gathered from credible sources, such as journals, books, national (expert) consensus documents, and official government websites. These factual sentences served as the premises in the dataset. The sentences then underwent a paraphrasing process. During this stage, each premise was duplicated multiple $n$ times and paraphrased to increase both the number and variation of premise sentences. Afterward, the premise sentences were processed by the hypothesis generator, where pairs of hypothesis sentences were generated. For each premise, multiple hypothesis sentences were generated, each labeled as entailment, contradiction, or neutral. Both the premise paraphraser and hypothesis generator processes used a zero-shot prompting technique. Table 3 describes the prompts used to generate the dataset. Finally, any possible duplicates were removed to ensure the uniqueness of the dataset.

**Table 3** Prompts used in the dataset generation workflow.

| Task | Prompt |
|---|---|
| Generate sentence pairs labeled as 'entailment'. | *Buatkan daftar (1,2,3,...)* $\{n\}$ *kalimat yang berhubungan dengan pernyataan '*$\{s\}$*' tidak lebih dari* $\{l\}$ *kata berbahasa Indonesia menggunakan EYD! Kalimat tidak mengandung unsur organisasi, politik, nama tokoh, dan SARA!* |
| Generate sentence pairs labeled as 'neutral'. | *Buatkan daftar (1,2,3,...)* $\{n\}$ *kalimat yang netral (tidak berhubungan) dengan pernyataan '*$\{s\}$*' tidak lebih dari* $\{l\}$ *kata berbahasa Indonesia menggunakan EYD! Kalimat tidak mengandung unsur organisasi, politik, nama tokoh, dan SARA!* |
| Generate sentence pairs labeled as 'contradiction'. | *Buatkan daftar (1,2,3,...)* $\{n\}$ *kalimat yang bertentangan dengan pernyataan '*$\{s\}$*' tidak lebih dari* $\{l\}$ *kata berbahasa Indonesia menggunakan EYD! Kalimat tidak mengandung unsur organisasi, politik, nama tokoh, dan SARA!* |

| Task | Prompt |
|---|---|
| Paraphrase a sentence. | *Parafrase menjadi kalimat berita untuk awam maksimal* $\{l\}$ *kata yang tidak boleh sama persis ataupun sebagian dengan hasil parafrase sebelumnya.: '*$\{s\}$*'* |

$\{s\}$: Input sentence; $\{n\}$: Number of generated sentences; $\{l\}$: Maximum length of the sentence.

To ensure the quality of the generated dataset, an evaluation focused on correctness was conducted. The primary goal was to verify that the generated sentence pairs matched the given labels. This evaluation was performed manually by two independent evaluators.

### 3.3 Experiment Design

The key focus of this study was the PLM, where the NLI and fact modules were replaced by the selected PLM. The experiment was designed to identify the PLM that resulted in the best performance compared to the baseline. The baseline referred to a model that did not use knowledge from a KG and was defined as a PLM directly connected to the classifier module. The PLMs evaluated included indolem/indobert [38] and indobenchmark/indobert (p1 and p2) [39] as monolingual models, as well as mBERT [40] and XLM-RoBERTa [41] as multilingual models. All PLMs included in this study were of the case-insensitive (uncased) type and based on the Transformer Base architecture. Meanwhile, the KG used in this study was COVID-19 KG Bahasa Indonesia [27].

The experiments were divided into two phases. The first phase trained the model and identified the best hyperparameter configuration, while the second phase was focused on testing the model. During the first phase, the model was trained using the training dataset, and validation was conducted using the validation dataset. In the second phase, testing was performed using the testing dataset. The models were trained with a learning rate of 2e-5, a batch size of 16, and 16 epochs, employing an early stopping strategy with a patience of 5. The loss function used was cross-entropy loss, and the optimizer was Adam. Training was conducted on an Intel Xeon Silver 4208 processor and an Nvidia Quadro RTX 5000 GPU with 16 GB of RAM. The evaluation metrics included precision, recall, accuracy, and F1-score. The Wilcoxon Signed-Rank test was used to assess the statistical significance of the resulting accuracy.

## 4 Result and Discussion

### 4.1 Generated Dataset

From our dataset generation workflow, we created 18,750 premise-hypothesis sentence pairs, with each label (entailment, contradiction, neutral) having 6,250 sentence pairs (Table 4). The dataset was then divided into training and testing datasets with a ratio of 80:20. The training set was further divided into training and validation datasets with a ratio of 80:20. Therefore, this resulted in a 64%, 16%, and 20% dataset division for training, validation, and testing, respectively. The use of the 80:20 split was based on the Pareto principle, which states that 80% of effects come from 20% of causes [42], and this strategy is commonly used in NLP experiments [43,44,45]. Evaluation of correctness was conducted on 100 randomly selected samples by two independent authors. The first evaluator gave a score of 90%, while the second evaluator gave a score of 87%. This resulted in an overall dataset correctness score of 88.5%, sufficient for this study.

**Table 4** Examples from the generated dataset.

| Sentence Pair | Label |
|---|---|
| **Premise:** *Protein RBD pada Spike Covid-19 berperan berinteraksi dengan sel tubuh secara langsung.* **Hypothesis:** *Fungsi RBD dalam Spike Covid- 19 adalah berhubungan langsung dengan sel tubuh.* | Entailment |
| **Premise:** *Obat Remdesivir melalui infus disetujui untuk mengobati COVID-19 pada orang dewasa dan anak- anak.* **Hypothesis:** *Obat Remdesivir yang diberikan melalui infus tidak direkomendasikan untuk mengobati COVID-19 orang dewasa dan anak-anak.* | Contradiction |
| **Premise:** *COVID-19 dapat menyebabkan peradangan yang meningkatkan kemungkinan terjadinya pembekuan darah.* **Hypothesis:** *Pencegahan penyebaran COVID-19 melibatkan mencuci tangan, menggunakan masker, dan menjaga jarak.* | Neutral |

## 4.2 Model Evaluation

Table 5 shows the results of the first phase experiment (the training phase). Although the model was run for 16 epochs, the experiments indicated that the model achieved the best results within the first 5 epochs. Most of the model's best results were obtained after just 2 training epochs. Among the models, the one using XLM-RoBERTa [41] required the longest training time, reaching its best performance at 5 epochs. This can possibly be explained by the fact that XLM-RoBERTa [41] had the largest number of parameters compared to the other models. From this, one can infer that an early stopping strategy can be used for an effective and efficient training process, reducing the need for longer epochs, which typically offer only marginal improvements and thus minimize the computer resources required.

**Table 5** Results of the first phase of the experiment (training). Only the best results are shown in this table.

| Model Architecture | Epoch | Loss | Precision | Recall | Accuracy | F1 |
|---|---|---|---|---|---|---|
| indolem/indobert [38] | | | | | | |
| Baseline | 3 | 0.3925 | 0.8573 | 0.8538 | 0.8553 | 0.8530 |
| Proposed | 2 | **0.3728** | **0.8610** | **0.8547** | **0.8557** | **0.8544** |
| indobenchmark/indobert p1 [39] | | | | | | |
| Baseline | 2 | 0.4349 | 0.8322 | 0.8307 | 0.8320 | 0.8310 |
| Proposed | 2 | **0.4370** | **0.8486** | **0.8328** | **0.8330** | **0.8335** |
| indobenchmark/indobert p2 [39] | | | | | | |
| Baseline | 2 | 0.4462 | 0.8243 | 0.8237 | 0.8253 | 0.8239 |
| Proposed | 2 | **0.4294** | **0.8405** | **0.8275** | **0.8277** | **0.8287** |
| mBERT [40] | | | | | | |
| Baseline | 2 | 0.4369 | 0.8254 | 0.8167 | 0.8170 | 0.8180 |
| Proposed | 2 | **0.4324** | **0.8306** | **0.8297** | **0.8313** | **0.8296** |
| XLM-RoBERTa [41] | | | | | | |
| Baseline | 5 | 0.4196 | 0.8466 | 0.8460 | 0.8480 | 0.8460 |
| Proposed | 5 | **0.3907** | **0.8609** | **0.8551** | **0.8560** | **0.8552** |

According to Table 5, it is evident that our proposed model architecture consistently yielded the best results across all evaluation metrics used compared to its baseline. This indicates that the use of a KG added valuable information to the model, enhancing its performance. The best performance was achieved by using XLM-RoBERTa [41] as the PLM, with an accuracy of 0.8560. Meanwhile, the lowest performance was exhibited by using indobenchmark/indobert p2 [39] as the PLM, with an accuracy of 0.8277.

To evaluate real-world performance, the best models for both the baseline and proposed approaches, as determined from the first phase of the experiment, were tested using the test dataset. Table 6 shows the results of the second phase of the experiment (the testing). From the table, it can be observed that our proposed model architecture consistently outperformed its baseline. Moreover, the use of the XLM-RoBERTa [41] PLM yielded the best result, with an accuracy of up to 0.8616. Compared to the baseline, the improvement resulted was 1.65%. In contrast, the model that used mBERT [40] as the PLM yielded the lowest result, with an accuracy as low as 0.8277. The Wilcoxon Signed-Rank Test further strengthened the significance of the XLM-RoBERTa's performance, with a $p$-value < 0.05.

**Table 6** Results of the second phase of the experiment (testing)

| Model Architecture | Precision | Recall | Accuracy | F1 | *p*-value |
|---|---|---|---|---|---|
| indolem/indobert [38] | | | | | |
| Baseline | 0.8576 | 0.8548 | 0.8555 | 0.8539 | 0.432 |
| Proposed | **0.8642** | **0.8590** | **0.8592** | **0.8588** | |
| indobenchmark/indobert p1 [39] | | | | | |
| Baseline | 0.8341 | 0.8330 | 0.8336 | 0.8334 | 0.31 |
| Proposed | **0.8512** | **0.8396** | **0.8395** | **0.8404** | |
| indobenchmark/indobert p2 [39] | | | | | |
| Baseline | 0.8363 | 0.8360 | 0.8368 | 0.8358 | 0.833 |
| Proposed | **0.8502** | **0.8373** | **0.8371** | **0.8390** | |
| mBERT [40] | | | | | |
| Baseline | 0.8331 | 0.8249 | 0.8248 | 0.8263 | 0.77 |
| Proposed | **0.8271** | **0.8270** | **0.8277** | **0.8268** | |
| XLM-RoBERTa [41] | | | | | |
| Baseline | 0.8443 | 0.8441 | 0.8451 | 0.8436 | **0.01*** |
| Proposed | **0.8654** | **0.8614** | **0.8616** | **0.8615** | |

*$p$-value < 0.05, statistically significant

Table 7 shows the number of true predictions across the PLMs used. From the table it can be inferred that, except for the use of mBERT, the use of a KG in our proposed model increased the number of entailment class predictions. This can be explained by the fact that the KG added valuable information to the model, which led to an increase in true entailment class predictions. However, despite the improvement, this came with a tradeoff, as the number of contradiction and neutral class predictions decreased.

**Table 7** Number of true predictions across PLMs used.

| PLM | Baseline | | | Experimental | | |
|---|---|---|---|---|---|---|
| | E | C | N | E | C | N |
| indolem/indobert [38] | 1038 | 1211 | 959 | 1087 | 1172 | 963 |
| indobenchmark/indobert p1 [39] | 971 | 1136 | 1019 | 1096 | 1103 | 949 |
| indobenchmark/indobert p2 [39] | 933 | 1148 | 1057 | 1089 | 1068 | 982 |
| mBERT [40] | 1038 | 1059 | 996 | 953 | 1145 | 1006 |
| XLM-RoBERTa [41] | 932 | 1183 | 1054 | 1082 | 1165 | 984 |

E: Entailment; C: Contradiction; N: Neutral

### 4.3 Error Analysis

Error analysis was performed to understand where the model still fell short. In this case, error analysis was performed on the XLM-RoBERTa, the best PLM used in our proposed model. Tables 8 and 9 show examples of the test dataset that were predicted correctly and incorrectly, respectively. From the tables, it can be observed that the model attempted to return the most relevant fact paragraph information given the hypothesis sentence. The relevant keywords are marked with underscores. This provided additional information for the model to make better predictions.

However, despite these improvements, one issue identified was the need for a better algorithm to return the relevant information for the given hypothesis sentence. Our word-matching level mechanism relied heavily on word-to-word matching and did not consider the surrounding context. This resulted in non-relevant fact paragraphs being returned. Furthermore, another issue arose when the information was not available in the KG, resulting in empty returned fact paragraphs. These limitations may have contributed to the best accuracy of the model being limited to 0.8616. Therefore, further research is needed to improve the information retrieval algorithm and the completeness of the KG for fact-checking purposes.

**Table 8** Examples from the test dataset that were predicted correctly.

| Data | Pred | Label |
|---|---|---|
| **Premise:** *Penelitian terbaru menemukan bahwa COVID-19 dapat terus menular melalui udara selama 3 jam.*<br>**Hypothesis:** *Penelitian terbaru menunjukkan bahwa COVID-19 dapat menular melalui tetesan udara selama 3 jam.*<br>**Fact paragraph:** *COVID-19 terdiri atas terkonfirmasi. COVID-19 ditularkan melalui droplet udara.* | E | E |
| **Premise:** *Infeksi virus saat hamil dapat meningkatkan risiko keguguran, kelahiran prematur, dan lahir mati.*<br>**Hypothesis:** *Konsultasikan dengan dokter untuk mengatasi risiko infeksi virus selama kehamilan.*<br>**Fact paragraph:** *COVID-19 terdiri atas reinfeksi. Severe Acute Respiratory Syndrome Coronavirus-2 (SARS-CoV-2) terdiri atas BA.5.* | N | N |
| **Premise:** *Varian omikron SARS-CoV-2 menurunkan efektivitas casirivimab dan imdevimab, berdasarkan bukti baru yang ditemukan.*<br>**Hypothesis:** *Varian omikron SARS-CoV-2 tidak berdampak pada efektivitas casirivimab dan imdevimab.*<br>**Fact paragraph:** *Severe Acute Respiratory Syndrome Coronavirus-2 (sars-cov-2) terdiri atas omicron.* | C | C |

**Table 9** Examples from the test dataset that were predicted incorrectly.

| Data | Pred | Label |
|---|---|---|
| **Premise:** *Ilmuwan berhasil menemukan virus SARS-CoV-2 dalam sampel jantung pasien yang terinfeksi.*<br>**Hypothesis:** *Virus SARS-CoV-2 berhasil diisolasi dari sampel jantung pasien yang terinfeksi.*<br>**Fact paragraph:** *Severe Acute Respiratory Syndrome Coronavirus-2 (SARS-CoV-2) terdiri atas delta. Severe Acute Respiratory Syndrome Coronavirus-2 (SARS-CoV-2) terdiri atas delta. COVID-19 memiliki komplikasi tamponade jantung.* | C | E |
| **Premise:** *Pasien tanpa gejala COVID-19 tidak mengalami perubahan yang signifikan pada sel darah dan peradangan.*<br>**Hypothesis:** *Pasien COVID-19 tanpa gejala tidak mengalami perubahan yang signifikan pada tingkat peradangan.*<br>**Fact paragraph:** *COVID-19 memiliki komplikasi anemia hemolitik autoimun.* | E | N |
| **Premise:** *Penelitian menyarankan agar bronkoskopi tidak digunakan pada pasien COVID-19 karena risiko penyebaran melalui udara.*<br>**Hypothesis:** *Beberapa studi menyarankan bronkoskopi tetap dapat dilakukan dengan langkah-langkah pencegahan yang tepat.*<br>*Fact paragraph: -* | N | C |

E: Entailment; N: Neutral; C: Contradiction

## 5 Conclusion

In this paper, we proposed using a KG to enhance NLI performance for automated COVID-19 fact-checking in Indonesian language. Our model processed semantic relationships between premise and hypothesis sentences and KG-derived information in separate modules, then combined their representation vectors as input to the classifier. This approach enabled the integration of semantic and KG-based information while keeping model complexity low. The best performance was achieved using XLM-RoBERTa, trained with a learning rate of 2e-5 for 5 epochs using cross-entropy loss and the Adam optimizer, yielding an accuracy of 0.8616, i.e., 1.65% higher than the baseline. However, error analysis revealed limitations. First, KG incompleteness may have reduced the available information, limiting fact retrieval. Second, the retrieval mechanism relied solely on keyword matching, ignoring contextual cues, which likely impacted both the quantity and accuracy of retrieved facts. Future work should focus on enriching the KG and developing more effective, context-aware retrieval mechanisms. In closing, despite our study focusing on COVID-19, our proposed model architecture can be used for other cases of automated fact-checking, increasing its accuracy through the use of a KG.

## Acknowledgements

This study was the first author's master's thesis, completed under the second author's supervision. The authors thank Farhan Hilmi Taufikulhakim, Khartika Mahardini, and Muhammad Taufik Suryono from the Faculty of Medicine, Universitas Indonesia, for their invaluable help in compiling COVID-19-related factual sentences for the dataset. The authors also appreciate Patrick Segara for refining the figures and extend special thanks to Harits Abdurrohman, Muhammad Anwari Leksono, and Nur Ahmadi for their invaluable discussions during the journal writing process. Lastly, the authors acknowledge the Informatics master's program at Institut Teknologi Bandung for providing their high-performance computing resources.

## Source of Funding

This work was supported by the Thesis Research Grant for a master's degree from the National Competitive Research Program Fund, Ministry of Education, Culture, Research, and Technology, Republic of Indonesia, in 2022.